\documentclass[runningheads]{llncs}
\usepackage{graphicx}
\usepackage[utf8]{inputenc}
\usepackage[english]{babel}
\usepackage{xcolor,colortbl}
\usepackage{blindtext}
\usepackage{caption}
\usepackage{subcaption}
\usepackage{booktabs}
\usepackage{multirow, makecell}
\usepackage{breakcites} 

\usepackage{hyperref}
\hypersetup{
    colorlinks=true,
    linkcolor=blue,
    filecolor=magenta,      
    urlcolor=magenta,
    pdftitle={Self-supervised Pre-training of Text Recognizers},
    pdfpagemode=FullScreen,
}

\usepackage{pifont}

\begin{document}
\title{Self-supervised Pre-training of Text Recognizers}
%
%
\author{Martin Kišš\inst{1} \orcidID{0000-0001-6853-0508} \and
Michal Hradiš\inst{1} \orcidID{0000-0002-6364-129X}}
\authorrunning{M. Kišš, M. Hradiš}
%
\institute{
Faculty of Information Technology, Brno University of Technology, \\
Brno, Czech Republic\\
\email{\{ikiss,hradis\}@fit.vutbr.cz}
}
\maketitle              
\begin{abstract}
In this paper, we investigate self-supervised pre-training methods for document text recognition. 
Nowadays, large unlabeled datasets can be collected for many research tasks, including text recognition, but it is costly to annotate them. 
Therefore, methods utilizing unlabeled data are researched.
We study self-supervised pre-training methods based on masked label prediction using three different approaches -- Feature Quantization, VQ-VAE, and Post-Quantized AE.
We also investigate joint-embedding approaches with VICReg and NT-Xent objectives, for which we propose an image shifting technique to prevent model collapse where it relies solely on positional encoding while completely ignoring the input image.
We perform our experiments on historical handwritten (Bentham) and historical printed datasets mainly to investigate the benefits of the self-supervised pre-training techniques with different amounts of annotated target domain data. 
We use transfer learning as strong baselines.
The evaluation shows that the self-supervised pre-training on data from the target domain is very effective, but it struggles to outperform transfer learning from closely related domains.
This paper is one of the first researches exploring self-supervised pre-training in document text recognition, and we believe that it will become a cornerstone for future research in this area. 
We made our implementation of the investigated methods publicly available at \url{https://github.com/DCGM/pero-pretraining}.

\end{abstract}
\section{Introduction}
Nowadays, large models are an integral part of machine learning.
Representatives of such models are, for example, language models GPT-3~\cite{brown_language_2020} or LLaMa-2~\cite{touvron_llama_2023}, the~ASR model Whisper~\cite{radford_robust_2022}, or generative models such as Stable Diffusion~\cite{rombach_high-resolution_2022} or DALLE-3~\cite{betker_improving_nodate}. 
In order for these models to generalize, a large amount of training data is needed. 
However, obtaining such large datasets with annotations is very expensive either in terms of cost or time. 
This also applies to text recognition and especially to handwritten text recognition.

At the same time, relatively large datasets without annotations are available and approaches based on semi-supervised or self-supervised learning are able to utilize this data. 
Semi-supervised approaches are mostly based on pseudo-labeling~\cite{lee_pseudo-label_2013,kiss_at-st_2021,kiss_softctcsemi-supervised_2023,xie_self-training_2020} (also self-training) or consistency regularization~\cite{berthelot_mixmatch_2019,kurakin_remixmatch_2020,aberdam_multimodal_2022}. 
Examples of self-supervised approaches include wav2vec 2.0~\cite{baevski_wav2vec_2020}, HuBERT~\cite{hsu_hubert_2021}, or BEST-RQ~\cite{chiu_self-supervised_2022} for ASR, and BEiT~\cite{bao_beit_2022,peng_beit_2022,wang_image_2023}, DINO~\cite{caron_emerging_2021,oquab_dinov2_2024}, iBOT~\cite{zhou_ibot_2022}, I-JEPA~\cite{assran_self-supervised_2023}, or VICReg~\cite{bardes_vicreg_2022} for image modeling. 
These approaches are often based on masked label prediction techniques or joint-embedding learning.
The self-supervised approaches are typically used to pre-train a model on a large unlabeled dataset and then the model is fine-tuned to the specific downstream task, such as speech recognition or image classification.
In the area of text recognition, semi-supervised methods have been investigated relatively intensively, but self-supervised methods have not been studied that much.
There are a few methods~\cite{aberdam_sequence--sequence_2021,yang_reading_2022,guan_self-supervised_2023} based on contrastive learning that is applied to scene text recognition task -- images that typically contain one or only a few words in natural environment (e.g. various signs on buildings, banners, etc).

This paper investigates self-supervised methods for text recognition on full text lines.
In particular, we investigate methods based on joint-embedding learning and masked label prediction.
The main contributions of the paper are as follows:
\begin{enumerate}
    \item We investigate the benefits of model pre-training using self-supervised methods on both printed and handwriting datasets in a challenging scenario with a limited number of annotated text lines and we compare them to strong transfer learning baselines.
    \item We propose a shifting technique for joint-embedding learning to prevent the models from relying solely on positional encoding while completely ignoring the input image.
    \item We make our implementation of the pre-training methods publicly available so anyone can follow our research and contribute to the research of the self-supervised methods in the document text recognition area.
\end{enumerate}

\section{Related work}
In recent years, self-supervised learning has emerged as a successful paradigm in machine learning for training models without relying on large-scale labeled datasets. 
In general, the model is pushed to learn meaningful representations through supervision signals generated from the input data.
Self-supervised learning has been successfully used in language modeling~\cite{devlin_bert_2018}, image modeling~\cite{bao_beit_2022,peng_beit_2022,wang_image_2023,caron_emerging_2021,oquab_dinov2_2024,zhou_ibot_2022,bardes_vicreg_2022} or automatic speech recognition (ASR)~\cite{baevski_wav2vec_2020,hsu_hubert_2021,chiu_self-supervised_2022,chen_wavlm_2022}.

Self-supervised learning often utilizes a masked label prediction technique where the input or intermediate representation is masked and the model is trained to classify the masked part based on its context.
Labels in this case are not the actual labels from the downstream task the model is later trained on, but they are inferred from the input data and it is desired that the same labels describe similar parts of the input data.
Using this approach, the model learns dependencies between the surrounding inputs and the target label, thereby learning to represent the input data.
The masking-based modeling approach was first used in BERT~\cite{devlin_bert_2018} pre-training of Transformer-based language models.

\subsection{ASR Methods}
One of the first approaches to self-supervised learning in ASR is wav2vec 2.0~\cite{baevski_wav2vec_2020} by Baevski et al.
In this approach, the raw audio is processed using a convolutional encoder producing a latent representation.
Parts of the latent representation are randomly masked and the entire sequence is further processed by Transformer Encoder.
The latent representation is also quantized using product quantization.
The model is then trained using a contrastive loss on the outputs corresponding to the masked inputs where the positive sample and the negative samples are selected from the quantized latent representation.

Another self-supervised approach is HuBERT~\cite{hsu_hubert_2021} by Hsu et al.
In HuBERT, the audio is processed by a convolutional encoder, masked, and further processed by a Transformer.
The training target labels are generated from the audio by calculating Mel-frequency cepstral coefficients (MFCCs) and clustered by the kMeans algorithm.
A similar approach to HuBERT is WavLM~\cite{chen_wavlm_2022}.
In the WavLM system, the input audio is also augmented with noise and/or other audio, allowing the model to also learn speaker recognition, separation, and diarization.
BEST-RQ~\cite{chiu_self-supervised_2022} is yet another self-supervised approach utilizing masked training.
In BEST-RQ, the labels are generated from the audio by a randomly initialized and frozen projection and codebook.

\subsection{Image Methods}
In image modeling, the self-supervised approaches are represented by DINO~\cite{caron_emerging_2021,oquab_dinov2_2024}, iBOT~\cite{zhou_ibot_2022}, SwAV~\cite{caron_unsupervised_2020}, BEiT~\cite{bao_beit_2022,peng_beit_2022,wang_image_2023}, SimCLR~\cite{chen_simple_2020}, and many others.
In BEiT~\cite{bao_beit_2022}, masked image modeling is used to pre-train a Transformer-based model which is then fine-tuned to image classification and semantic segmentation.
Labels for the masked training are produced using Discrete Variational AutoEncoder (dVAE) which is trained beforehand to reconstruct images.
In later versions (BEiT v2~\cite{peng_beit_2022} and BEiT-3~\cite{wang_image_2023}) the dVAE is replaced with the VQ-KD model.
In DiT~\cite{li_dit_2022}, a similar approach to the BEiT is used to pre-train a model on document images.

Some approaches are based on joint-embedding learning, such as VICReg~\cite{bardes_vicreg_2022} or SimCLR~\cite{chen_simple_2020}.
In VICReg (Variance-Invariance-Covariance Regularization), the proposed loss function pushes the model to produce the same output embeddings for two views of the same input image while retaining variance in the outputs and decorrelating the individual values in the output.
In an extension called VICRegL~\cite{bardes_vicregl_2022}, the same criterion is also applied to local patches.
NT-Xent (Normalized Temperature-scaled Cross Entropy)~\cite{chen_simple_2020} is a loss function based on standard cross entropy, modifications of which have been successfully used in various approaches such as SimCLR~\cite{chen_simple_2020} or CLIP~\cite{radford_learning_2021}.
The loss operates on two $l_2$-normalized sequences of vectors pulling the similarity of corresponding vectors towards 1 and similarity to the rest of the vectors to 0.

\subsection{Text Recognition}
The self-supervised pre-training methods in text recognition are mainly focused on contrastive learning for scene text recognition~\cite{aberdam_sequence--sequence_2021,yang_reading_2022,guan_self-supervised_2023} and learning degradation invariant models~\cite{souibgui_text-diae_2023}.
In SeqCLR~\cite{aberdam_sequence--sequence_2021}, two augmented views of a word image are processed by a model trained using noise contrastive estimation loss function.
Similarly, in DiG~\cite{yang_reading_2022} approach proposed by Yang et al., the two views of a word image are processed by a model trained using InfoNCE loss function.
Moreover, they add a reconstruction head to the model which aims to reconstruct the masked original image.
CCD~\cite{guan_self-supervised_2023} approach is similar to the previous two approaches but it adds character-based segmentation which helps the contrastive loss function to better estimate the corresponding outputs when geometric augmentations are applied.
Text-DIAE~\cite{souibgui_text-diae_2023} is an approach where an AutoEncoder model is trained to reconstruct visually degraded input images.
Then the encoder part is directly trained to recognize handwritten or scene text.

\section{Self-supervised Pre-training of Text Recognizers}
\begin{figure}[t]
    \begin{subfigure}{0.305\linewidth}
        \centering
        \includegraphics[width=\linewidth]{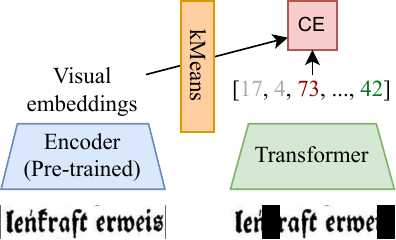}
        \caption{Feature Quantization}
        \label{fig:method_fq}
    \end{subfigure}
    \hfill
    \begin{subfigure}{0.277\linewidth}
        \centering
        \includegraphics[width=\linewidth]{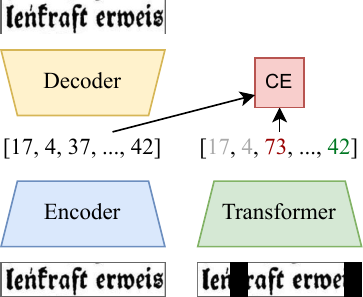}
        \caption{VQ-VAE}
        \label{fig:method_vqvae}
    \end{subfigure}
    \hfill
    \begin{subfigure}{0.305\linewidth}
        \centering
        \includegraphics[width=\linewidth]{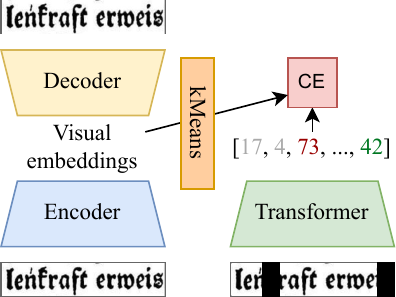}
        \caption{Post-Quantized AE}
        \label{fig:method_pqae}
    \end{subfigure}
    \vspace{0.5cm} \\ 
    \begin{subfigure}{0.362\linewidth}
        \centering
        \includegraphics[width=\linewidth]{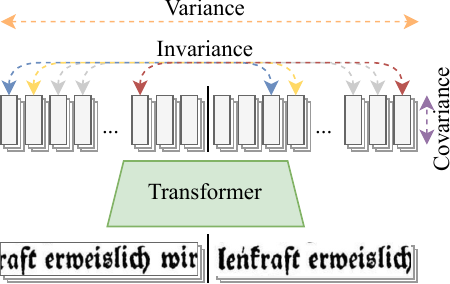}
        \caption{VICReg}
        \label{fig:method_vicreg}
    \end{subfigure}
    \hfill
    \begin{subfigure}{0.466\linewidth}
        \centering
        \includegraphics[width=\linewidth]{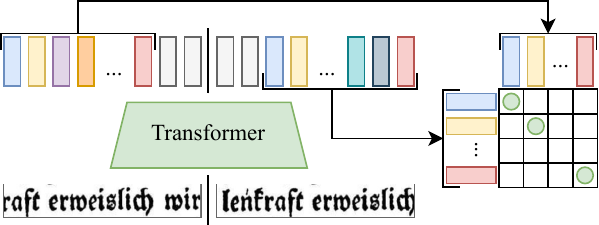}
        \caption{NT-Xent}
        \label{fig:method_ntxent}
    \end{subfigure}
    \caption{Overview of the investigated self-supervised pre-training approaches. 
    The first row of images depicts methods based on predicting a label of a masked input.
    The second row shows joint-embedding methods.
    }
    \label{fig:methods}
\end{figure}

Our goal is to train well-performing text recognition models without excessive human annotation effort.
Thus, we investigate the possibilities of self-supervised pre-training methods of such models on unlabeled data.
We divide the methods into two categories: \emph{masked label prediction}, where parts of the input data are masked and the model predicts the correct label, and \emph{joint-embedding methods} utilizing different loss functions.
Figure~\ref{fig:methods} shows the overview of the investigated self-supervised pre-training approaches, which are described in detail in the following paragraphs.

\subsection{Masked label prediction}
We investigate several methods which follow similar principles of predicting masked labels.
Having input data (e.g. image, audio, etc.) and a set of corresponding labels, it is possible to train the model to predict the labels at masked locations.
The crucial question here is how to obtain these labels.
We explore three methods how to get them: Feature Quantization using kMeans algorithm, Vector Quantised-Variational AutoEncoder (VQ-VAE), and combination of the previous two which we refer to as Post-Quantized AutoEncoder.

\paragraph{Feature Quantization}
The Feature Quantization (see Figure~\ref{fig:method_fq}) method is inspired by HuBERT~\cite{hsu_hubert_2021} used in ASR where the labels are obtained by clustering MFCCs of the input audio using the kMeans algorithm.
In our case, we use a convolutional encoder to obtain visual features which are then clustered using kMeans.
While it is possible to use generally any encoder, we use encoders from models trained on text recognition, thereby leveraging their domain knowledge.

\paragraph{VQ-VAE}
The second method (see Figure~\ref{fig:method_vqvae}) is inspired by BEiT~\cite{bao_beit_2022} and DiT~\cite{li_dit_2022} where the labels are generated using the Discrete Variational AutoEncoder (dVAE) model trained to reconstruct images and documents, respectively.
In our case, we use a VQ-VAE model which is very similar to dVAE.
The model is trained to reconstruct images of text lines and thus it should encode visually similar parts into similar discrete values.
Labels for the masked pre-training are generated using the encoder and the quantization mechanism.

\paragraph{Post-Quantized AE}
The Post-Quantized AE method combines the previous two (see Figure~\ref{fig:method_pqae}).
The main problem we experienced when training the VQ-VAE model was inconsistency of the training which is heavily dependent on the initialization of the quantization mechanism.
Therefore, we propose to use the basic AE model without quantization and to do the quantization after the training using the kMeans algorithm as in the Feature Quantization approach.

\subsection{Joint-embedding methods}
\label{sec:joint-embedding_methods}
To train our models with joint-embedding methods, we take two differently augmented views of the same text line image, process them by the model, and calculate a loss function on the two sequences of outputs.
We train our models using two different loss functions -- VICReg~\cite{bardes_vicreg_2022} and NT-Xent~\cite{chen_simple_2020}.
Both loss functions train the model to produce distinctive outputs that are close for visually similar parts of the input.
Where these two loss functions fundamentally differ is that while VICReg is computed over the entire batch, NT-Xent is computed separately within each line.

\paragraph{VICReg}
The VICReg~\cite{bardes_vicreg_2022} (see Figure~\ref{fig:method_vicreg}) is a non-contrastive joint-embedding criterion consists of three terms: (1) a variance term, which ensures that the individual outputs are variable enough, (2) an invariance term, which minimizes the distance between the corresponding outputs, and (3) a covariance term decorrelating the values in the outputs.

\paragraph{NT-Xent}
The NT-Xent~\cite{chen_simple_2020} (normalized temperature-scaled cross-entropy; see Figure~\ref{fig:method_ntxent}) is a contrastive loss function which maximizes the agreement between the corresponding outputs while minimizing the agreement between the non-corresponding outputs.
The agreement is calculated using a similarity metric and a temperature, which is a hyperparameter.

\begin{figure}[t]
    \begin{subfigure}{0.47\linewidth}
        \centering
        \includegraphics[width=\linewidth]{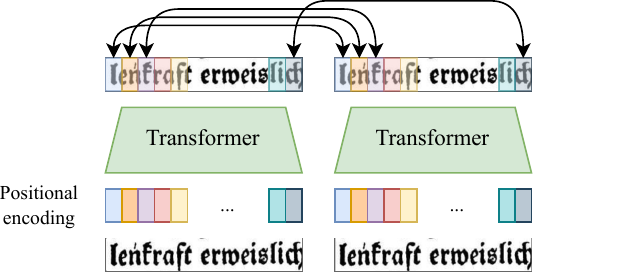}
        \caption{Without shifting}
        \label{fig:without_shifting}
    \end{subfigure}
    \hfill
    \begin{subfigure}{0.47\linewidth}
        \centering
        \includegraphics[width=\linewidth]{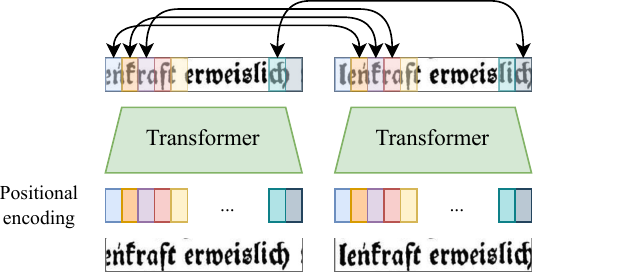}
        \caption{With shifting}
        \label{fig:with_shifting}
    \end{subfigure}
    \caption{Illustration of the proposed shifting technique. If shifting is applied (right image), different positional encoding is added to the corresponding parts of the two input images. Without shifting (left image), the same positional encoding is added, resulting in training collapse.}
    \label{fig:shifting}
\end{figure}

\paragraph{Image shifting}
When training a model on text lines with the described loss functions, the training can collapse quite easily.
Since we add positional encoding to the image representations inside the model, it can simply learn to appropriately transform the positional information to minimize both objectives, completely ignoring the input image.
To overcome this issue, we propose to randomly horizontally shift the two views of a text line.
This ensures that a different positional encoding is added to the corresponding parts of the input image, forcing the model to use information from the input image.
For illustration, see Figure~\ref{fig:shifting}.

\section{Experiments}
\label{sec:experiments}
We experimentally demonstrate the benefits of the investigated pre-training methods on handwritten and printed text recognition.
We follow the traditional self-supervised scheme where we first pre-train the model using one of our pre-training methods on all available training data and then we fine-tune the model using standard supervised training.
We fine-tune the models on three different amounts of annotated text lines (100, 1k, and 10k) to evaluate the performance of the pre-training methods in challenging low-resource scenarios.
For the evaluation, we use the standard character error rate (CER) metric.
We compare the fine-tuned models to models trained from scratch on the respective training datasets using a standard supervised training approach.
Moreover, we compare them also to models trained first on related text recognition datasets and then fine-tuned on the respective target datasets, i.e. transfer learning.
Besides the CERs, we also show reconstructions produced by the VQ-VAE and AE models and visualizations of similarly or equally encoded parts of text line images using procedures described later in this section.

\subsection{Datasets}
\label{sec:datasets}
\begin{figure}[t]
    \begin{subfigure}{0.48\linewidth}
        \centering
        \includegraphics[width=\linewidth]{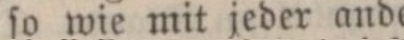}
        \caption{DTA dataset}
        \label{fig:datasets_dta}
    \end{subfigure}
    \hfill
    \begin{subfigure}{0.48\linewidth}
        \centering
        \includegraphics[width=\linewidth]{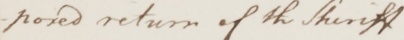}
        \caption{Bentham dataset}
        \label{fig:datasets_bentham}
    \end{subfigure}
    \\
    \begin{subfigure}{0.48\linewidth}
        \centering
        \includegraphics[width=\linewidth]{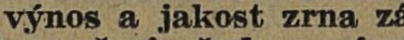}
        \caption{PERO-PRINTED}
        \label{fig:datasets_pero-printed}
    \end{subfigure}
    \hfill
    \begin{subfigure}{0.48\linewidth}
        \centering
        \includegraphics[width=\linewidth]{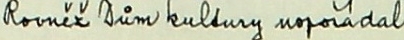}
        \caption{PERO-HWR}
        \label{fig:datasets_pero-hwr}
    \end{subfigure}
    \\
    \begin{subfigure}{\linewidth}
        \centering
        \includegraphics[width=0.48\linewidth]{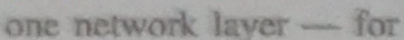}
        \caption{B-MOD}
        \label{fig:datasets_b-mod}
    \end{subfigure}
    \caption{Examples of text lines from datasets.}
    \label{fig:datasets}
\end{figure}

We conduct the experiments on both handwritten and printed text recognition tasks. Statistics of all datasets are summarized in Table~\ref{tab:datasets} and examples are shown in Figure~\ref{fig:datasets}.

As the main handwriting datasets, we selected the ICFHR 2014 Bentham Dataset~\cite{sanchez_icfhr2014_2014} and pages collected from the Bentham Project\footnote{\url{https://www.ucl.ac.uk/bentham-project}}.
Both the ICFHR dataset and the pages collected online are written in historical English and they are written by only a few authors, but only the ICFHR dataset contains line-level annotations.
Note also that the training set of the ICFHR Bentham dataset contains only 9k text lines, but for simplicity, we refer to it as 10k in the results.
We used our layout detector~\cite{kodym_page_2021} based on the ParseNet architecture to obtain text lines on the unlabeled pages.
Our main printed text dataset is a set of pages collected from Deutsches Textarchiv\footnote{\url{https://www.deutschestextarchiv.de/}} which we refer to as the DTA dataset.
Each page of this collection contains line-level annotations.
This dataset is written mostly in historical German with Fraktur as the dominant font.

\begin{table}[t]
    \centering
    \caption{%
        Sizes of the used datasets in the number of lines.
    }\label{tab:datasets}
    \begin{tabular}{
        @{\extracolsep{4pt}}@{\kern\tabcolsep}
        lrrr}
        \toprule    

        Dataset
        & \multicolumn{1}{c}{Training}
        & \multicolumn{1}{c}{Validation}
        & \multicolumn{1}{c}{Test}
        \\
        
        \midrule
        DTA            & 2\,593\,813 & 24\,699 & 25\,824 \\ \addlinespace[0.1cm]
        Bentham \\
        -- annotated   &      9\,194 &  1\,415 &     860 \\
        -- unannotated & 1\,141\,566 &      -- &      -- \\ \addlinespace[0.1cm]
        B-MOD          &    403\,744 & 52\,091 & 59\,565 \\ \addlinespace[0.1cm]
        PERO-PRINTED   & 1\,825\,977 &  2\,814 &  2\,814 \\ \addlinespace[0.1cm]
        PERO-HWR       &    362\,199 &  3\,700 &  3\,700 \\

        \bottomrule
    \end{tabular}
\end{table}

For the transfer learning experiments, we use other various datasets which are more or less related to the main two.
The first one is B-MOD~\cite{kiss_brno_2019} which contains a large collection of pages from mostly scientific articles captured by mobile devices.
The other two are PERO-PRINTED and PERO-HWR which we collected using our online OCR application PERO-OCR\footnote{\url{https://pero-ocr.fit.vutbr.cz/}} and both contain various printed and handwritten pages, respectively.

\subsection{Models}
\label{sec:models}
In our experiments, we use two model architectures, both based on Transformers.
The first one is the standard Vision Transformer (ViT), which consists of a patch embedding layer and a series of Transformer Encoder blocks with Multi-Head Self-Attention~\cite{vaswani_attention_2017,dosovitskiy_image_2021}.
On top of this backbone, the architecture contains a head, which architecture depends on the training approach used, but it is always a single linear layer or MLP. 
The second architecture contains a VGG-based convolutional encoder instead of the patch embedding layer.
We refer to this architecture as VggT.

In particular, our models operate on images of text lines with a normalized height of 40 pixels.
The patch embedding layer in the ViT architecture takes $40\times8$ slices of the input image and encodes them into a sequence of feature vectors which are then processed by the Transformer.
Similarly, the VGG encoder of the VggT architecture transforms the input image into a sequence of feature vectors with a horizontal subsampling factor of $8$, i.e. the sequence is  $8\times$ shorter in the horizontal dimension than the original input.
The VGG encoder\footnote{We use weights from the VGG16 model which is pre-trained on ImageNet and is available in \texttt{torchvision.models.vgg16}.} contains 10 convolutional layers with ReLU activation function, max-pooling layers, and batch normalization.
Both architectures contain six Transformer Encoder layers with dimension 512, 8 attention heads, and MLP expansion ratio 4.
We also use sine and cosine-based absolute positional encoding which is added to the feature vectors.

\subsection{Training details}
\label{sec:training}

\begin{table}[t]
    \centering
    \caption{%
        Training setups.
    }\label{tab:training}
    \begin{tabular}{
        @{\extracolsep{4pt}}@{\kern\tabcolsep}
        lrccccc}
        \toprule
        Method & Iterations & Warm-up & Learning rate & Batch size & Head & Aug. \\        
        \midrule

        \multicolumn{7}{l}{\textit{Masked label prediction}} \\ \addlinespace[0.1cm]
        &    0--300k & 10k & $2\times 10^{-4}$ & 32 & Linear & \\
        & 300k--500k &     & $1\times 10^{-4}$ & 48 & Linear & \\
        & 500k--700k &     & $5\times 10^{-5}$ & 48 & Linear & \\
        \midrule

        \multicolumn{7}{l}{\textit{VQ-VAE, AE}} \\ \addlinespace[0.1cm]
        &    0--100k & 10k & $2\times 10^{-4}$ &  64 & -- & Visual \\
        & 100k--250k &     & $1\times 10^{-4}$ & 128 & -- & Visual \\
        & 250k--300k &     & $5\times 10^{-5}$ & 128 & -- & Visual \\
        \midrule
        
        \multicolumn{7}{l}{\textit{VICReg}} \\ \addlinespace[0.1cm]
        & 0--120k & 25k & $1\times 10^{-4}$ & 256 & MLP(3, 2048) & Visual \\
        \midrule
        
        \multicolumn{7}{l}{\textit{NT-Xent}} \\ \addlinespace[0.1cm]
        &     0--30k & 25k & $1\times 10^{-4}$ & 256 & Linear(2048) & Visual \\
        &  30k--100k &     & $5\times 10^{-5}$ & 256 & Linear(2048) & Visual \\
        & 100k--120k &     & $5\times 10^{-5}$ & 256 & Linear(2048) & Visual \\
        \midrule
        
        \multicolumn{7}{l}{\textit{OCR from scratch}} \\ \addlinespace[0.1cm]
        &    0--150k & 10k & $2\times 10^{-4}$ & 32 & Linear(512) & All \\ 
        & 150k--250k &     & $1\times 10^{-4}$ & 48 & Linear(512) & All \\
        & 250k--300k &     & $5\times 10^{-5}$ & 64 & Linear(512) & All \\
        \midrule

        \multicolumn{7}{l}{\textit{OCR fine-tuning}} \\ \addlinespace[0.1cm]
        &   0--10k & 10k & $2\times 10^{-4}$ & 32 & Linear(512) & All \\ 
        & 10k--40k &     & $1\times 10^{-4}$ & 48 & Linear(512) & All \\ 
        & 40k--50k &     & $5\times 10^{-5}$ & 64 & Linear(512) & All \\      

        \bottomrule
    \end{tabular}
\end{table}

As mentioned in the introduction of this section, we either pre-train our models using one of the investigated methods and then fine-tune them on the text recognition task or we train the models on the text recognition task -- directly on the target dataset or using transfer learning from related dataset.
We use Adam optimizer with training setups summarized in Table~\ref{tab:training}.
The table shows the settings of the number of iterations, warm-up iterations, learning rates, batch sizes, model heads, and augmentations used in the different trainings.
Model heads are denoted by the type and size of the layers.
The only parameter for the Linear head is the output size, which in the case of the masked label prediction depends on the underlying label generation approach and thus on the number of recognized classes.
In the case of MLP, the first parameter indicates the number of layers and the second parameter is their size.
In experiments involving VICReg criterion, we followed the original paper and thus the head is MLP.
After the pre-training phase, we replace the head with a new head which is designed for the OCR fine-tuning phase.
The \emph{Visual} augmentations consist of colour changing, adding noise, gamma adjusting, and motion and defocus blurring.
The \emph{All} augmentations comprise the Visual ones and also image geometry changing (e.g. skewing, scaling, etc.) and masking~\cite{kiss_at-st_2021}.

\subsubsection{Pre-training phase}
\paragraph{Feature Quantization}
Having a pre-trained encoder, we produce visual features of each line in the used dataset.
In our experiments, we use a convolutional encoder from an OCR model which was previously trained on different data.
Since the DTA and Bentham datasets are quite large, we sample $k \times 100$ random feature vectors from the respective datasets for the kMeans optimization.
In our setup, $k$ is the number of recognized classes and we use $k=4096$.
The optimization of the kMeans model is performed for 100 epochs with a batch size of 16k samples\footnote{We use the \texttt{MiniBatchKMeans} implementation from \texttt{sklearn.cluster}}.
Once we have the optimized kMeans model, we generate labels for the entire dataset.

\paragraph{Masked label prediction}
Using the generated labels, we train our models to predict the labels of the masked inputs.
We set the masking probability to $p=0.2$ and we mask slices of size $40\times8$ pixels of the input image with random noise.
The model is trained using the cross-entropy loss function.

\paragraph{AutoEncoders}
Our AutoEncoder models are based on the VGG architecture -- the encoder consists of 8 convolutional layers with ReLU activation function.
The decoder architecture is the same, just mirrored, and max-pooling layers are replaced by upsampling ones.
In the case of the VQ-VAE model, the model contains a Vector Quantization mechanism between the encoder and the decoder with a codebook of size 1\,024 when trained on the printed dataset and 2\,048 when trained on the handwriting dataset.
These codebook sizes showed the most promising results in our preliminary experiments.
Labels for the subsequent masked label prediction training are generated using the trained encoder and the Vector Quantization mechanism.
In the case of the Post-Quantized AutoEncoder approach, we generate labels from visual features produced by the trained encoder using the same scheme as in the Feature Quantization approach.
We train the AutoEncoder models using MSE loss.

\paragraph{Joint-embedding methods}
When we train the model using joint-embedding methods, we create random crops from the two augmented views of each sample.
This helps us create batches with higher variability in terms of the relative shifts between the two views while maintaining reasonable image widths.
From the first view we make the crop at random position and from the second view the crop is shifted to the right or left randomly with a step equal to the horizontal subsampling factor.
We also restrict that the two crops overlap by at least one horizontal subsampling factor.

We set the crop width to 512 and 256 pixels for the first 100k iterations on the printed and handwritten datasets, respectively.
For the rest of the training, we double the crop width.
We use the different sizes of crops on the two datasets mainly because the lines in the handwriting dataset are on average shorter.
For the NT-Xent loss function, we use a constant temperature of 0.1 throughout the entire training.

\subsubsection{Training the OCR}
We train our OCR models using CTC loss function~\cite{graves_connectionist_nodate}.
Model head is in this case a linear layer with a size of 512 which corresponds to the size of the recognized character set.
Data augmentation used during OCR training includes masking which we believe should improve the implicit language modeling capabilities of the model as proposed in AT-ST~\cite{kiss_at-st_2021} by Kišš et al.

\subsection{Results}
\label{sec:results}
In the conducted experiments, we compare the self-supervised pre-training approaches with baseline approaches on the printed and handwriting datasets.
We evaluate the trained models in terms of CER, we report error rates for the approaches based on the masked label prediction and we show reconstructed images produced by the trained AE-based models.

\paragraph{Visualizations}
All of the investigated pre-training approaches are based on the idea that visually similar parts of the input image should be encoded into similar labels or embeddings.
Thus, we want to visualize the similarly encoded image parts before or during the pre-training phase.

In the case of the joint-embedding methods, we have two batches of differently augmented views of text lines.
We generate outputs using the currently optimized model, we select one random output from each line in the first batch and we search for $N$ most similar outputs in the entire second batch.
Finally, we map the outputs back into the original images and display the corresponding parts side-by-side with a little context.

In the case of masked label prediction methods, selecting the same labels in the entire dataset and mapping them to the original image does not work.
We believe this is due to the fact that the labels are highly contextualized, i.e. parts of the image are not described by just one label, but they depend very much on the surrounding labels.
Therefore, we calculate trigrams of the labels and we search for image parts with the same label triplets in the entire dataset.

\subsubsection{Pre-training experiments}
\begin{figure}[t]
    \includegraphics[width=0.48\linewidth]{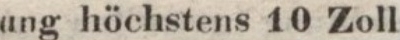}
    \hfill
    \includegraphics[width=0.48\linewidth]{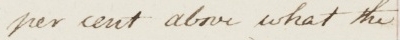}
    \\
    \includegraphics[width=0.48\linewidth]{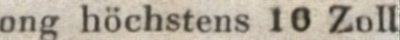}
    \hfill
    \includegraphics[width=0.48\linewidth]{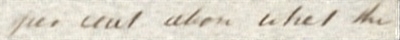}
    \\
    \includegraphics[width=0.48\linewidth]{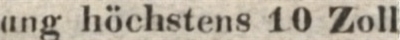}
    \hfill
    \includegraphics[width=0.48\linewidth]{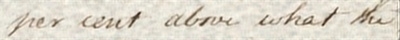}
    \caption{Examples of reconstructions using trained VQ-VAE and AE models on the DTA dataset (left column) and the Bentham dataset (right column). The first row shows the original images, the second and the third rows show reconstructions produced by VQ-VAE and AE models respectively. More reconstructions can be found in the GitHub repository.}
    \label{fig:reconstructions}
\end{figure}

\begin{figure}[t]
    \begin{subfigure}{0.48\linewidth}
        \centering
        \includegraphics[width=\linewidth]{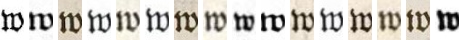}
        \caption{FQ (PERO-HWR)}
    \end{subfigure}
    \hfill
    \begin{subfigure}{0.48\linewidth}
        \centering
        \includegraphics[width=\linewidth]{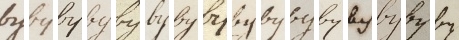}
        \caption{FQ (PERO-HWR)}
    \end{subfigure}
    \\
    \begin{subfigure}{0.48\linewidth}
        \centering
        \includegraphics[width=\linewidth]{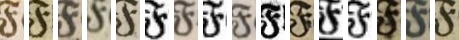}
        \caption{FQ (B-MOD)}
    \end{subfigure}
    \hfill
    \begin{subfigure}{0.48\linewidth}
        \centering
        \includegraphics[width=\linewidth]{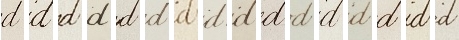}
        \caption{FQ (PERO-PRINTED)}
    \end{subfigure}
    \\
    \begin{subfigure}{0.48\linewidth}
        \centering
        \includegraphics[width=\linewidth]{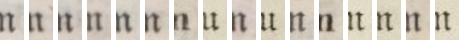}
        \caption{VQ-VAE}
    \end{subfigure}
    \hfill
    \begin{subfigure}{0.48\linewidth}
        \centering
        \includegraphics[width=\linewidth]{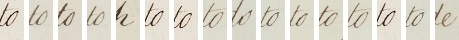}
        \caption{VQ-VAE}
    \end{subfigure}
    \\
    \begin{subfigure}{0.48\linewidth}
        \centering
        \includegraphics[width=\linewidth]{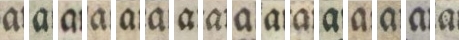}
        \caption{Post-Quantized AE}
    \end{subfigure}
    \hfill
    \begin{subfigure}{0.48\linewidth}
        \centering
        \includegraphics[width=\linewidth]{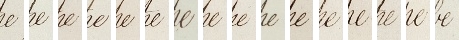}
        \caption{Post-Quantized AE}
    \end{subfigure}
    
    \caption{Examples of visually similar parts of images on the DTA dataset (left column) and the Bentham dataset (right column) extracted using quantization-based approaches. In figures (a)--(d) FQ stands for Feature Quantization and the dataset name in the parentheses indicates the dataset used to train the model from which the convolutional encoder is taken. More visualizations can be found in the GitHub repository.}
    \label{fig:visualizations_masked_pretraining}
\end{figure}

\paragraph{Masked label prediction}
Feature Quantization experiments are performed using convolutional encoders from models trained on PERO-HWR and B-MOD when pre-training on the DTA dataset and from models trained on PERO-HWR and PERO-PRINTED when pre-training on the Bentham dataset.
In Figure~\ref{fig:reconstructions} we show reconstructions produced by the VQ-VAE and AE models on validation sets and in Figure~\ref{fig:visualizations_masked_pretraining} we show the visualizations of label triplets as described earlier.
During the masked pre-training, we also measure the error rates of label prediction and we report them for the best-performing models in Table~\ref{tab:error_rates}.
In Table~\ref{tab:final} we report the CERs of the models fine-tuned on the text recognition task.
The results with really high CERs (around 70 \%) occur in cases when the model fails to generalize -- it learns the training set perfectly but it is not able to perform on the validation and test sets.

\begin{table}[t]
    \centering
    \caption{%
        Masked label prediction validation error rates. We report top-1, top-3, and top-10 error rates [\%]  on the validation sets of the respective datasets of the pre-training approaches utilizing masked label prediction. FQ stands for Feature Quantization and the dataset name in the parentheses indicates the dataset used to train the model from which the convolutional encoder is taken.
    }\label{tab:error_rates}
    \begin{tabular}{
        @{\extracolsep{4pt}}@{\kern\tabcolsep}
        llrrrrrr}
        \toprule
        \multirowcell{2}{Dataset} & 
        \multirowcell{2}{Method} &  
        \multicolumn{3}{c}{VggT} &
        \multicolumn{3}{c}{ViT} \\
        \cline{3-5}
        \cline{6-8}
        \addlinespace[0.1cm]
        & & \multicolumn{1}{c}{top-1} & \multicolumn{1}{c}{top-3} & \multicolumn{1}{c}{top-10} & \multicolumn{1}{c}{top-1} & \multicolumn{1}{c}{top-3} & \multicolumn{1}{c}{top-10} \\
        \midrule
        \multirowcell{4}{DTA} 
        & FQ (PERO-HWR)        & 26.31 & 5.56 & 1.07 & 28.87 & 6.83 & 1.27 \\
        & FQ (B-MOD)           & 30.86 & 8.26 & 1.58 & 33.59 & 10.04 & 2.17 \\
        & VQ-VAE               & 42.91 & 14.99 & 3.33 & 45.21 & 16.83 & 4.07 \\
        & Post-Quantized AE    & 32.73 &  9.81 & 1.96 & 33.86 & 10.59 & 2.20 \\
        
        \midrule
        \multirowcell{4}{Bentham\\dataset} 
        & FQ (PERO-HWR)        & 45.37 & 19.84 &  7.17 & 53.24 & 27.68 & 11.73 \\
        & FQ (PERO-PRINTED)    & 64.43 & 39.42 & 17.91 & 70.55 & 47.70 & 24.80 \\
        & VQ-VAE               & 63.50 & 38.15 & 16.59 & 68.86 & 44.95 & 21.43 \\
        & Post-Quantized AE    & 59.08 & 34.72 & 15.78 & 65.38 & 42.52 & 21.28 \\
        
        \bottomrule
    \end{tabular}
\end{table}

The error rates measured during masked pre-training are relatively high in some cases, especially on the Bentham dataset.
One reason for this may be that the number of recognized classes is too large, which may cause their overlapping and the trained model is not able to distinguish between them.
This is supported by the decreasing trend between the top-1, top-3, and top-10 error rates.
The other factor is that the handwritten dataset is significantly more difficult than the printed one.

\begin{figure}[t!]
    \begin{subfigure}{\linewidth}
        \centering
        \includegraphics[width=0.48\linewidth]{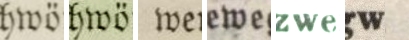}
        \hfill
        \includegraphics[width=0.48\linewidth]{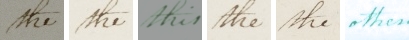}
        \caption{VICReg, VggT}
    \end{subfigure}
    \\
    \begin{subfigure}{\linewidth}
        \centering
        \includegraphics[width=0.48\linewidth]{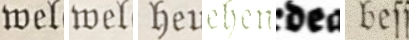}
        \hfill
        \includegraphics[width=0.48\linewidth]{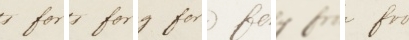}
        \caption{VICReg, ViT}
    \end{subfigure}
    \\
    \begin{subfigure}{\linewidth}
        \centering
        \includegraphics[width=0.48\linewidth]{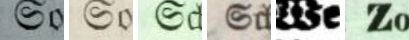}
        \hfill
        \includegraphics[width=0.48\linewidth]{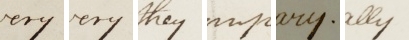}
        \caption{NT-Xent, VggT}
    \end{subfigure}
    \\
    \begin{subfigure}{\linewidth}
        \centering
        \includegraphics[width=0.48\linewidth]{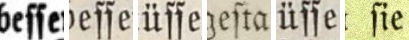}
        \hfill
        \includegraphics[width=0.48\linewidth]{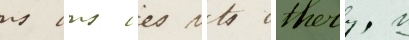}
        \caption{NT-Xent, ViT}
    \end{subfigure}
    \caption{Examples of visually similar parts of images on the DTA dataset (left column) and the Bentham dataset (right column) extracted during training by VggT and ViT models trained using VICReg and NT-Xent approaches. The first crop in the sequences is from the first batch and used as a query and the rest of the crops are the most visually similar parts in the second batch. More visualizations can be found in the GitHub repository.}
    \label{fig:visualizations_vicreg_ntxent}
\end{figure}

\paragraph{Joint-embedding methods}
Visualizations of the similarly encoded parts of input images are produced during the training of the model.
In Figure~\ref{fig:visualizations_vicreg_ntxent} we show visualizations obtained from the best performing models of both architectures, trained using both loss functions, and on both main datasets.
The CERs of the fine-tuned OCR models are in Table~\ref{tab:final}.

The proposed image shifting technique plays a crucial role when pre-training the models.
Without the shifting, models learn to simply transform the positional encoding into constant vectors (different for each position) satisfying the training criterion and neglecting the image information entirely.
When the shifting is applied as described in~\ref{sec:joint-embedding_methods}, the corresponding parts of the images are located at different positions and thus the trained model is forced to extract the visual information from inputs.

\subsubsection{Baselines}
Our baselines are OCRs trained from scratch and using transfer learning.
In Table~\ref{tab:final} we report CERs on the test sets of the DTA and Bentham datasets.
In transfer learning experiments, we first train the models on related datasets, where one is closer to the target dataset and the second is more distant.
For the DTA dataset, the closer one is the PERO-HWR dataset and the more distant one is the B-MOD.
The PERO-HWR dataset is closer in this case because, unlike B-MOD, it contains historical data and also handwritten text pages.
For the Bentham dataset, the closer related dataset is again the PERO-HWR dataset and the distant one is the PERO-PRINTED dataset.

\begin{table}[t!]
    \centering
    \caption{%
        Results of the baselines and the pre-training methods on test sets in CER [\%]. The values in bold represent the best results within the respective parts. Column headings 100, 1k, and 10k refer to the size of the used training dataset. TL and FQ stand for Transfer Learning and Feature Quantization, respectively. The dataset name in the parentheses indicates the dataset used to train the model from which the convolutional encoder is taken.
    }\label{tab:final}
    \begin{tabular}{
        @{\extracolsep{4pt}}@{\kern\tabcolsep}
        llrrrrrr}
        \toprule
        \multirowcell{2}{Dataset} & 
        \multirowcell{2}{Method} &  
        \multicolumn{3}{c}{VggT} &
        \multicolumn{3}{c}{ViT} \\
        \cline{3-5}
        \cline{6-8}
        \addlinespace[0.1cm]
        & & \multicolumn{1}{c}{100} & \multicolumn{1}{c}{1k} & \multicolumn{1}{c}{10k} & \multicolumn{1}{c}{100} & \multicolumn{1}{c}{1k} & \multicolumn{1}{c}{10k} \\
        \midrule
        \multirowcell{9}{DTA} 
        & OCR from scratch     & 5.26 & 1.30 & 0.53 & 72.83 & 70.57 & 0.75 \\
        & TL from B-MOD        & 6.20 & 1.50 & 0.60 & 7.96 & 2.89 & 1.15 \\
        & TL from PERO-PRINTED & \textbf{2.03} & \textbf{0.79} & \textbf{0.51} & \textbf{2.07} & \textbf{1.03} & \textbf{0.59} \\ \cline{2-8} \addlinespace[0.1cm]
        & FQ (PERO-HWR)        & \textbf{2.34} & \textbf{0.83} & \textbf{0.50} & \textbf{2.74} & \textbf{0.97} & \textbf{0.58} \\
        & FQ (B-MOD)           & 2.98 & 0.98 & 0.55 & 3.22 & 1.15 & 0.62 \\
        & VQ-VAE               & 2.99 & 0.95 & 0.52 & 3.70 & 1.29 & 0.65 \\
        & Post-Quantized AE    & 4.72 & 1.05 & 0.54 & 77.04 & 1.38 & 0.65 \\
        & VICReg               & 2.82 & 1.03 & 0.54 & 7.14 & 1.61 & 0.70 \\
        & NT-Xent              & 5.60 & 1.36 & 0.59 & 11.60 & 2.59 & 0.87 \\
        
        \midrule
        \multirowcell{9}{Bentham\\dataset} 
        & OCR from scratch     & 23.36 & 8.33 & 2.75 & 74.62 & 72.52 & 7.50 \\
        & TL from PERO-HWR     & \textbf{5.87} & \textbf{3.81} & \textbf{2.27} & \textbf{12.21} & \textbf{8.63} & 4.96 \\
        & TL from PERO-PRINTED & 18.78 & 6.90 & 3.11 & 23.51 & 10.14 & \textbf{4.68} \\ \cline{2-8} \addlinespace[0.1cm]
        & FQ (PERO-HWR)        & \textbf{10.09} & \textbf{4.80} & \textbf{2.91} & \textbf{17.66} & \textbf{8.13} & \textbf{4.75} \\
        & FQ (PERO-PRINTED)    & 17.46 & 7.25 & 3.70 & 23.91 & 11.05 & 5.47 \\
        & VQ-VAE               & 21.49 & 8.73 & 4.71 & 33.63 & 14.32 & 7.24 \\
        & Post-Quantized AE    & 23.13 & 8.86 & 4.72 & 76.86 & 15.09 & 7.45 \\
        & VICReg               & 19.58 & 8.35 & 3.60 & 27.38 & 10.92 & 4.96 \\
        & NT-Xent              & 30.58 & 9.48 & 3.60 & 57.54 & 22.71 & 7.05 \\
        
        \bottomrule
    \end{tabular}
\end{table}

\subsubsection{Results evaluation} 

From the measured CERs we can see that models trained using transfer learning on closely related datasets outperform the rest of the methods.
For the more distant related dataset, the results slightly favour the Feature Quantization pre-training method, especially on the printed dataset.
From the methods that do not require any previously trained OCR model, the VICReg and masked pre-training using VQ-VAE perform the best.
It is also worth noticing that even though the Post-Quantized AE method outperformed the VQ-VAE method in terms of error rates of the masked label prediction in the pre-training, VQ-VAE achieved better CERs after fine-tuning to text recognition.

For comparison, we trained also OCRs from scratch on the entire training set of the DTA dataset resulting in CERs of 0.28 \% and 0.81 \% on the test set by the VggT and ViT models respectively.
Regarding the Bentham dataset, the current state-of-the-art result on the test set is 3.30 \% CER~\cite{vidal_end-end_2023}.

\section{Conclusions}
In this paper, we investigated several self-supervised methods for pre-training text recognition models.
Specifically, we investigated approaches based on joint-embedding learning -- VICReg and NT-Xent -- and several methods for masked label prediction -- Feature Quantization, VQ-VAE, and Post-Quantized AE.
We also proposed a technique that prevents models trained with joint-embedding methods from learning to simply transform the positional encoding without considering the input image at all.
We evaluated the pre-training approaches on the printed DTA dataset and the handwriting Bentham dataset.
We first pre-trained a model using a given approach and then we fine-tuned the model to text recognition on three portions of text lines (100, 1k, 10k lines) of the target training dataset.
Our baselines are OCRs trained from scratch on the target dataset and models trained using transfer learning from related datasets.

The obtained results show that the best strategy is to use transfer learning from closely related datasets, but this may not always be possible.
Among the pre-training methods, Feature Quantization, which requires a pre-trained encoder, performed the best.
From the approaches that do not require any pre-trained model, VICReg and masked pre-training with VQ-VAE showed the best results.
However, the differences in the character error rates show that there is a huge performance gap worth researching, especially on the handwritten dataset.
We believe that this paper will start extensive research in the self-supervised pre-training approaches to text recognition.
To support that, we published our implementation of the pre-training methods online in GitHub repository\footnote{\url{https://github.com/DCGM/pero-pretraining}}.

\subsubsection{Acknowledgement}
This work has been supported by the Ministry of Culture Czech
Republic in NAKI III project semANT - Semantic Document Exploration
(DH23P03OVV060).

%
%
%
\bibliographystyle{splncs04}
\bibliography{mybib}

\end{document}